\documentclass{article}
\usepackage[T1]{fontenc}
\usepackage[utf8]{inputenc}
\usepackage{amsmath,amssymb,amsthm,booktabs,tabularx,graphicx,natbib}
\usepackage{tikz,pgfplots}
\usetikzlibrary{arrows.meta,positioning}
\pgfplotsset{compat=1.18}
\usepackage{hyperref}
\usepackage{url}
\hypersetup{colorlinks=true,linkcolor=blue,citecolor=blue,urlcolor=blue,
  pdftitle={XU-RS: Explaining Credal Width in Random-Set Language Models},
  pdfauthor={}}
\newcommand{\xurs}{XU-RS}
\newcommand{\Bel}{\operatorname{Bel}}
\newcommand{\Pl}{\operatorname{Pl}}
\newcommand{\BetP}{\operatorname{BetP}}
\newcommand{\EG}{\operatorname{EG}}

\newcommand{\E}{\mathbb{E}}
\newcommand{\R}{\mathbb{R}}

\newcommand{\EffectRows}{%
SmolLM3-3B & 0.0381 & [0.0243, 0.0532]\\
Llama-2-7B & 0.0450 & [0.0295, 0.0617]\\
}

\newcommand{\CorrectionRows}{%
SmolLM3: head only & 95.84 & 1.2311 \\
SmolLM3: partial tuning & 99.84 & 2.1623 \\
SmolLM3: LoRA & 99.92 & 1.7221 \\
Llama-2: LoRA, seed 7 & 99.14 & 2.2889 \\
Llama-2: LoRA, seed 17 & 99.61 & 2.4980 \\
Llama-2: LoRA, seed 23 & 99.92 & 2.8108 \\
}

\newcommand{\NumericsRows}{%
Training prompts & 0.9951 & 0.0652 & 0.1563 \\
Aligned paired prompt & 1.0000 & 0.0144 & 0.0384 \\
Zero-embedding baseline & 0.9840 & 0.6428 & 0.3375 \\
}

\newcommand{\EndpointRows}{%
family007\_case00 & $-0.478592$ & $+0.002640$ & 0.481233 & $1.10\times10^{-5}$ \\
family094\_case02 & $-0.538493$ & $-0.164970$ & 0.373522 & $4.61\times10^{-7}$ \\
family098\_case00 & $-0.629888$ & $-0.427342$ & 0.202545 & $7.39\times10^{-6}$ \\
family017\_case00 & $-0.619253$ & $-0.082523$ & 0.536730 & $6.51\times10^{-6}$ \\
}

\title{XU-RS: Explaining Credal Width in\\Random-Set Language Models}
\author{David Achara, Maryam Sultana, Alexander D. Rast, Fabio Cuzzolin \\
Institute for AI, Data Analysis and Systems (AIDAS) \\
Oxford Brookes University}
\begin{document}
\maketitle
\begin{abstract}
Uncertainty estimates tell us how unsure a model is, but not why. Without knowing which parts of an input influences a model's uncertainty, we cannot tell whether that uncertainty score depends on input features that are relevant for the task. We study this problem in random-set classifiers built using pretrained language models. These classifiers assign probability to individual answers and to groups of answers, producing lower and upper probabilities for each answer; The difference between these probabilities, called credal width, is used to represent epistemic uncertainty about an answer arising from limited training data. We propose XU-RS, a framework that attributes an answer's credal width to the input tokens (words or word pieces) supplied to a language model. XU-RS uses Expected Gradients (a standard feature attribution method) to estimate how input tokens contribute to credal width. The proposed framework is evaluated on a MedQA dataset using SmolLM3-3B and Llama-2-7B models, demonstrating that setting the embedding of a token ranked highly by XU-RS to zero (zero-masking) causes larger changes in credal width than zero-masking randomly selected tokens. In addition, we show that normalisation can cause other answer groups to influence an answer's width, reveal how token attribution can mask numerical errors, and provide diagnostic checks to verify whether a token ranked highly by XU-RS meaningfully explains model uncertainty.
\end{abstract}
\section{Introduction}

Uncertainty quantification \cite{kong2023uncertainty,cuzzolin2021uncertainty} accompanies a model's predictions with
estimates of their uncertainty \citep{huellermeier2021uncertainty}, which can be distinguished into either \emph{aleatoric} or \emph{epistemic} \cite{Swiler09epistemicuncertainty,oren2022mcts,li_2025_esi} uncertainty \cite{walters2023investor,bickfordsmith_2025_rethinking}.
These estimates do not, by themselves, explain which parts of the input
influence them. Understanding this dependence is important when examining
whether the features influencing uncertainty are actually meaningful for the task.
We investigate this problem for \emph{random-set} \cite{matheron1975random,molchanov2005theory,nguyen2006introduction,cuzzolin2023reasoning} classifiers built using
pretrained language models.

Consider a medical question with four answer options: $\{A,B,C,D\}$. A conventional
classifier assigns a probability to each option. This specifies how probability is
divided between the options, but does not by itself express how certain the model is about that division. A random-set classifier \cite{manchingal2022epistemic}
can instead assign probability jointly to a set of options \cite{wang2026set}, leaving
its division among them unspecified
\citep{manchingal2023random,manchingal2025rsnn,mubashar2025rsllm}. This flexibility is intended
to represent incomplete knowledge about the answer probabilities, and requires novel ways of assessing performance \cite{manchingal2025unifiedevaluationframeworkepistemic}.

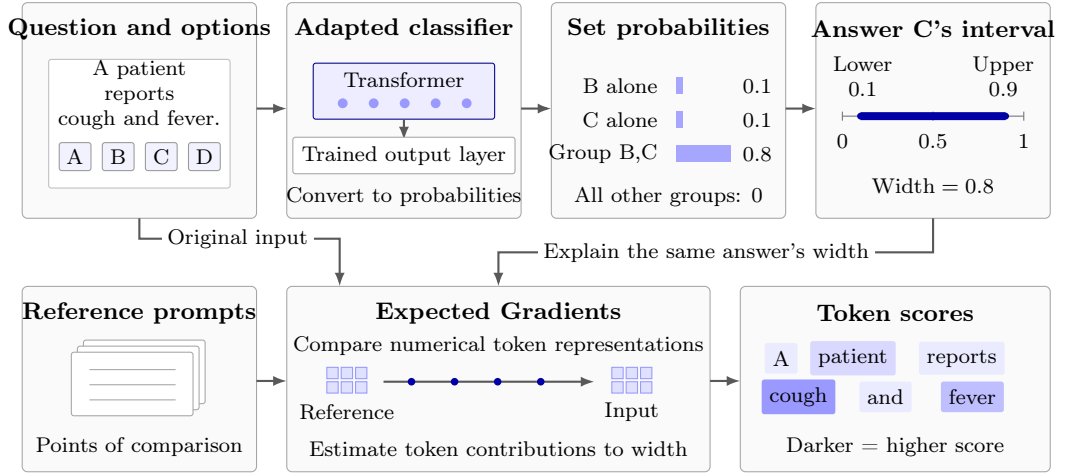
\begin{figure}[t]
  \centering
  \begin{tikzpicture}[
  x=1cm,y=1cm,
  font=\small,
  panel/.style={
    draw=black!35,
    rounded corners=2pt,
    fill=black!2
  },
  heading/.style={
    font=\small\bfseries,
    align=center
  },
  detail/.style={
    font=\footnotesize,
    align=center
  },
  arrow/.style={
    -{Latex[length=1.8mm]},
    thick,
    draw=black!65
  },
  token/.style={
    rounded corners=1pt,
    inner sep=3pt,
    font=\footnotesize
  }
]

% Forward computation.
\foreach \x in {0,3.5,7,10.5}
  \draw[panel] (\x,3.35) rectangle +(3.1,2.85);

\node[heading] at (1.55,5.85) {Question and options};
\node[heading] at (5.05,5.85) {Adapted classifier};
\node[heading] at (8.55,5.85) {Set probabilities};
\node[heading] at (12.05,5.85) {Answer C's interval};

% Question card and answer options.
\draw[draw=black!25,fill=white,rounded corners=1pt]
  (0.35,3.75) rectangle (2.75,5.45);

\node[detail,text width=2.2cm] at (1.55,4.98)
  {A patient reports\\cough and fever.};

\foreach \label/\x in {A/0.7,B/1.27,C/1.84,D/2.41} {
  \draw[draw=black!40,fill=blue!5,rounded corners=1pt]
    (\x-0.21,3.98) rectangle (\x+0.21,4.35);
  \node[detail] at (\x,4.165) {\label};
}

% Transformer and trained output layer.
\draw[draw=blue!55!black,fill=blue!7,rounded corners=1pt]
  (3.85,4.66) rectangle (6.25,5.4);

\node[detail] at (5.05,5.18) {Transformer};

\foreach \x in {4.24,4.65,5.06,5.47,5.88}
  \fill[blue!40] (\x,4.87) circle (0.065);

\draw[arrow] (5.05,4.66)--(5.05,4.38);

\node[
  detail,
  draw=black!35,
  fill=white,
  rounded corners=1pt,
  minimum width=2.4cm,
  minimum height=0.37cm
] at (5.05,4.18) {Trained output layer};

\node[detail] at (5.05,3.63) {Convert to probabilities};

% Illustrative probabilities; bars share the same scale.
\foreach \label/\y/\value in {
  B alone/5.1/0.1,
  C alone/4.65/0.1,
  {Group B,C}/4.2/0.8
} {
  \node[detail,anchor=east] at (8.53,\y) {\label};
  \fill[blue!35]
    (8.65,\y-0.11) rectangle ++({0.9*\value},0.22);
  \node[detail,anchor=west] at (9.41,\y) {\value};
}

\node[detail] at (8.55,3.63) {All other groups: 0};

% Probability interval for answer C.
\draw[draw=black!55] (10.85,4.7)--(13.25,4.7);

\foreach \x/\label in {10.85/0,12.05/0.5,13.25/1} {
  \draw[draw=black!55] (\x,4.64)--(\x,4.76);
  \node[detail,anchor=north] at (\x,4.59) {\label};
}

\draw[draw=blue!65!black,line width=3pt]
  (11.09,4.7)--(13.01,4.7);

\fill[blue!65!black] (11.09,4.7) circle (0.045);
\fill[blue!65!black] (13.01,4.7) circle (0.045);

\node[detail] at (11.12,5.23) {Lower\\0.1};
\node[detail] at (12.98,5.23) {Upper\\0.9};
\node[detail] at (12.05,3.78) {$\text{Width}=0.8$};

\draw[arrow] (3.1,4.8)--(3.5,4.8);
\draw[arrow] (6.6,4.8)--(7,4.8);
\draw[arrow] (10.1,4.8)--(10.5,4.8);

% Attribution computation.
\draw[panel] (0,0) rectangle (3.1,2.45);
\draw[panel] (3.5,0) rectangle (9.1,2.45);
\draw[panel] (9.5,0) rectangle (13.6,2.45);

\node[heading] at (1.55,2.1) {Reference prompts};
\node[heading] at (6.3,2.1) {Expected Gradients};
\node[heading] at (11.55,2.1) {Token scores};

% Reference-prompt cards.
\foreach \dx/\dy in {0.16/0.16,0.08/0.08,0/0} {
  \draw[draw=black!35,fill=white,rounded corners=1pt]
    (0.65+\dx,0.74+\dy) rectangle (2.26+\dx,1.58+\dy);
}

\foreach \y in {0.95,1.15,1.35}
  \draw[draw=black!35] (0.89,\y)--(2.03,\y);

\node[detail] at (1.55,0.33) {Points of comparison};

% Numerical representations and sampled comparison points.
\foreach \x/\label in {4.03/Reference,7.8/Input} {
  \foreach \i in {0,1,2}
    \foreach \j in {0,1}
      \draw[draw=blue!40,fill=blue!12]
        (\x+0.19*\i,1.04+0.19*\j) rectangle ++(0.15,0.15);

  \node[detail] at (\x+0.265,0.79) {\label};
}

\draw[arrow] (4.78,1.19)--(7.57,1.19);

\foreach \x in {5.15,5.72,6.29,6.86}
  \fill[blue!65!black] (\x,1.19) circle (0.05);

\node[detail] at (6.3,1.66)
  {Compare numerical token representations};

\node[detail] at (6.3,0.3)
  {Estimate token contributions to width};

% Illustrative scores, not measured attributions.
\node[token,fill=blue!8]  at (10.04,1.5) {A};
\node[token,fill=blue!15] at (10.99,1.5) {patient};
\node[token,fill=blue!8]  at (12.43,1.5) {reports};

\node[token,fill=blue!40] at (10.27,0.99) {cough};
\node[token,fill=blue!8]  at (11.42,0.99) {and};
\node[token,fill=blue!25] at (12.57,0.99) {fever};

\node[detail] at (11.55,0.33) {Darker = higher score};

% Inputs to the explanation and its output.
\draw[arrow] (3.1,1.2)--(3.5,1.2);
\draw[arrow] (9.1,1.2)--(9.5,1.2);

\draw[arrow]
  (1.55,3.35)--(1.55,3.08)--(4.2,3.08)--(4.2,2.45);

\node[detail,fill=white,inner sep=2pt]
  at (2.85,3.08) {Original input};

\draw[arrow]
  (12.05,3.35)--(12.05,2.9)--(6.3,2.9)--(6.3,2.45);

\node[detail,fill=white,inner sep=2pt]
  at (9.05,2.9) {Explain the same answer's width};

\end{tikzpicture}
  \caption{\xurs{} predicts answer intervals and attributes a selected
answer's width to input tokens relative to reference prompts.
The example assigns C interval $[0.1,0.9]$ and width 0.8.}
  \label{fig:pipeline-intro}
\end{figure}

For example, suppose the random-set classifier assigns
probability 0.1 to B alone, 0.1 to C alone, and 0.8 to the set containing
B and C, with zero assigned to all other sets. For answer C, 0.1 is
committed to that answer alone, while another 0.8 could belong to either
B or C. C's lower probability is therefore 0.1 and its upper probability
is 0.9. The difference, $0.9-0.1=0.8$, is called \emph{credal width}.
The random-set classifier measures the probability left uncommitted between C and its alternatives
(Figure~\ref{fig:pipeline-intro}).

Two models can report the same width while responding to different
features: one to conflicting evidence and another to the order of the
answer options. We therefore ask: \emph{which parts of a question and
its answer options influence the width reported for a selected answer?}
The explanation target is uncertainty, rather than the score used to
choose an answer.

We introduce \xurs{} to explain the widths produced by these classifiers. We evaluate \xurs{} on four-option random set classifiers adapted
from SmolLM3-3B and Llama-2-7B
\citep{smollm3,touvron2023llama2}. A trained output layer estimates
lower probabilities for answer sets. These estimates are converted
into probabilities assigned to the sets, from which the answer
intervals are calculated. The models therefore classify answer options
rather than generate free-form text.

\xurs{} uses Expected Gradients \citep{erion2021expected}, a standard
feature-attribution method, to assign scores to input tokens (words or word pieces). The method explains how an output changes from a
specified starting input to the input being studied. This starting
input is called a reference, or baseline: it defines the difference
that the token contributions are intended to explain.

In our MedQA experiments, reference prompts contain other questions
and answer options. We explain the difference between the selected
answer's width on the original input and its average width on the
references supplied to the calculation. Expected Gradients gradually
changes the references' numerical token representations towards those
of the original prompt, measuring how the selected answer's width
responds along the way.
Section~\ref{sec:framework} explains how these references are prepared.

Our contributions are threefold. First, we develop and test a framework
for attributing credal width to input tokens. On 100 MedQA questions
per model \citep{jin2021medqa}, setting the embedding of tokens ranked highest by XU-RS (\emph{zero-masking}) produces larger absolute width
changes than zero-masking randomly selected tokens. Second, we derive
how converting classifier outputs into valid probabilities makes an
answer's width depend on values assigned to sets excluding that answer.
Third, we provide numerical and reference-preparation checks: controlled
examples show that stable rankings can conceal inaccurate attribution
values, and that preparing a reference can alter the width difference
even when attribution accurately accounts for that altered difference.
\section{Related work}

\paragraph{Random-set and credal prediction.}
Random-set models, with the approach called `epistemic AI' (\cite{cuzzolin2024epistemic,manchingal2024epistemic,manchingal2025epistemic}), allow probability to be assigned to sets of possible
outcomes without specifying how it is divided among their members
\citep{shafer1976mathematical,cuzzolin2021geometry}.
Random-Set Neural Networks (RS-NN) implement this representation for
classification, including text classification
\citep{manchingal2025rsnn}.
Random-Set Large Language Models (RS-LLM) extend it to language generation,
predicting over sets of possible next tokens
\citep{mubashar2025rsllm}.
Other approaches obtain lower and upper probabilities without predicting
probabilities for answer sets, for instance those using credal sets, i.e., convex sets of distributions \cite{levi1980enterprise,caprio2025credal}. Credal Deep Ensembles combine networks
trained to predict probability bounds \citep{wang2024credalensembles,wang2026learning,wang2026credal},
while the Credal Wrapper constructs bounds from the predictions of
Bayesian neural networks or deep ensembles
\citep{wang2025credalwrapper}, using the intersection probability (\cite{cuzzolin2009credal,cuzzolin2007properties}) as pointwise estimator. Credal INNs \cite{wang2025creinns} combine the expressive power of interval neural networks with credal representations.
For language models, \citet{yang2026imprecise} investigate prompting
models to report lower and upper probabilities for possible answers,
as well as constructing intervals from multiple reported predictions.
The credal approach has been recently extended to large language models \cite{manchingal2026credal}, generative adversarial networks (\cite{mubashar2026epistemic}) and to a wrapper applicable to Bayesian NN (\cite{sultana2025epistemic}).
Our classifiers follow the random-set approach of RS-NN and RS-LLM (\cite{mubashar2025rsllm}),
but predict over four answer options rather than vocabulary tokens.
Second-order probability representation has been applied in the past to pose estimation (\cite{gong2017belief}), tracking (\cite{cuzzolin1999evidential}), logistic regression (\cite{cuzzolin2018belief}), clustering (\cite{zhu2026tdcc}), feature fusion (\cite{pang2025guest}), max-entropy classification (\cite{cuzzolin2018generalised}), statistical learning theory (\cite{cuzzolin2024generalising,caprio2024credal}) and a variety of other tasks (\cite{liu2019evidence,faza2026direct,manchingal2025uncertainty}).
A recent review can be found in \cite{wang2025review}.
We investigate how input tokens influence the resulting credal width,
rather than proposing a new uncertainty representation.

\paragraph{Uncertainty attribution.}
Previous work explains uncertainty either by assigning contributions
to input features or by finding changes to an input that reduce the
model's uncertainty.
CLUE searches for similar inputs on which the model is less uncertain,
using a generative model to keep the changes plausible
\citep{antoran2021clue}.
\citet{wang2025optimization} jointly learn which image regions to modify
and how to modify them to reduce uncertainty.
These approaches explain uncertainty through proposed changes to the
input.

Other methods assign uncertainty contributions directly to input
features.
\citet{perez2022uncertainty} combine comparisons with less uncertain
inputs and gradient-based attribution.
\citet{wang2023uncertainty} trace uncertainty estimates in Bayesian
neural networks back to individual image pixels.
\citet{watson2023uncertainty} use Shapley values to allocate changes
in conditional entropy (the uncertainty remaining when specified
features are known) among input features.
\citet{iversen2025drivers} apply existing attribution methods to
predicted variance, targeting aleatoric uncertainty associated with
variability in the data.
\xurs{} follows this general strategy of explaining an uncertainty
output, specialising it to the credal width of a selected answer option.
This differs from assessing uncertainty in an explanation:
\citet{marx2023uncertainty} construct uncertainty sets for feature
explanations, whereas our target is the uncertainty estimate reported
by the classifier.

\paragraph{Gradient-based token attribution.}
Integrated Gradients (IG) is a feature-attribution method that estimates
how input features contribute to a difference in a model's score
\citep{sundararajan2017axiomatic}.
It compares the original input with a chosen comparison input, called
a reference or baseline. For text, this reference can be another prompt.
For example, if a classifier reports an answer's width as 0.6 on the
original prompt and 0.2 on the reference prompt, the attribution
explains the difference of 0.4.

To perform this comparison, the method works with the numerical vectors
that represent the tokens, rather than changing the text directly.
It evaluates intermediate vectors between the reference and the
original input. Halfway between them, for example, each vector's values
are the averages of the corresponding reference and input values.
At these intermediate points, IG measures how small changes to the
vectors affect the chosen score. These sensitivities are called
gradients. The method combines them with the differences between the
original and reference vectors to estimate each feature's contribution.

Expected Gradients (EG) extends this approach by averaging contributions
over sampled references and intermediate vectors
\citep{erion2021expected}.
In \xurs{}, the references are reference prompts/comparison inputs (other questions and their answer options), and the score being
explained is the credal width of a selected answer option.

The reference matters because it determines what difference is being
explained \citep{sturmfels2020baselines}.
The intermediate vectors also require care: an average of two token
vectors need not represent an actual word.
Discretized Integrated Gradients therefore chooses intermediate
vectors that remain closer to the representations of actual words
\citep{sanyal2021discretized}.

\paragraph{Evaluation of attributions.}
An explanation that appears convincing to a reader might not
describe the model's full internal behaviour \citep{jacovi2020faithfulness}.
For example, \citet{adebayo2018sanity} test whether explanations respond
to changes in model parameters and training labels, showing why visual
inspection alone is insufficient.
Feature-replacement tests provide another form of evaluation: they
modify highly ranked features and measure the resulting change in
the model output.
However, the modified input may differ substantially from inputs
encountered during training. The measured change can therefore reflect
the model's response to an unfamiliar replacement, rather than the
importance of the original feature under ordinary inputs
\citep{hase2021ood}.

These studies motivate testing whether an attribution reflects the
model's behaviour, rather than judging it only by whether it looks
plausible. For \xurs{}, we distinguish evidence that tokens affect
credal width from evidence that they identify information meaningful
for the task.
\section{The XU-RS attribution framework}
\label{sec:framework}

\subsection{Random-set predictions and credal width}

Let $\Theta=\{A,B,C,D\}$ contain the answer options. A random-set
prediction assigns probability, called \emph{mass}, $m(S)$ to each
set $S\subseteq\Theta$, with $m(S)\geq0$, $m(\varnothing)=0$ and
$\sum_Sm(S)=1$. Mass on a set leaves its division among members
unspecified. Lower probability counts masses assigned entirely within
an event $E$; upper probability counts masses on sets overlapping $E$:
\begin{equation}
 \Bel(E)=\sum_{S\subseteq E}m(S),\qquad
 \Pl(E)=\sum_{S\cap E\ne\varnothing}m(S).
 \label{eq:belpl}
\end{equation}
These quantities are called belief and plausibility functions
\citep{shafer1976mathematical,cuzzolin2014belief,cuzzolin2018visions,cuzzolin2021geometry,cuzzolin2026statistical}.
For an answer $c$, credal width and the pignistic answer probability are
\begin{equation}
 W_c=\Pl(\{c\})-\Bel(\{c\})
     =\sum_{\substack{S\ni c\\|S|>1}}m(S),\qquad
 \BetP(c)=\sum_{S\ni c}\frac{m(S)}{|S|}.
 \label{eq:width}
\end{equation}
Here $|S|$ is the number of options in a set. Width lies in $[0,1]$
and counts probability assigned to sets containing $c$ together with
other options. 
A geometric approach to belief functions and other uncertainty measures can be found in \cite{cuzzolin2004geometry,cuzzolin2008geometric,cuzzolin2008credal,cuzzolin2020geometry-dempster}; their combinatorial analysis in \cite{cuzzolin2010three}, their algebraic properties in \cite{cuzzolin01bcc,cuzzolin2001lattice,zhou2017total}.
Selecting an answer requires a separate rule: the
pignistic probability divides each set's probability equally among
its members and adds the shares for each option. This division is used
to choose an answer; the original set probabilities still determine
its width.

We select $c=\arg\max_k\BetP(k)$ on the original prompt and explain
that same option throughout attribution and zero-masking. If modifying
the input changes the model's preferred answer, we still measure the
width of $c$. Otherwise, the measured difference would concern two
different answer options.

\subsection{Classifier architecture}

A linear output layer reads the transformer's representation at the
last non-padding token. Its 14 sigmoid outputs estimate lower
probabilities for four single-option sets, six pairs and four triples,
following the belief-output approach of RS-NN and RS-LLM
\citep{manchingal2025rsnn,mubashar2025rsllm}.

An estimated lower probability for a set is not the probability assigned
to that set alone. For example,
$\Bel(\{B,C\})=m(\{B\})+m(\{C\})+m(\{B,C\})$.
Recovering $m(\{B,C\})$ therefore requires subtracting the probabilities
already assigned to B and C individually. The implementation processes
single-option sets, pairs and triples in that order, subtracting only
the nonnegative parts of previously computed values for smaller sets.
Appendix~\ref{app:proofs} gives the exact recursion.

Although the sigmoid places each predicted lower probability in $[0,1]$,
it does not enforce the relationships between these predictions in
Equation~\ref{eq:belpl}. The conversion can consequently produce
negative values or a total different from one. A final correction
sets negative values to zero, assigns any shortfall to the set of all
four options and rescales the resulting values to sum to one.
Section~\ref{sec:correction} derives how this operation affects width.
The classifier uses the corrected probabilities during both training
and attribution.

\subsection{Reference prompts and Expected Gradients}

\paragraph{Preparing reference prompts.}
The transformer represents each token by an embedding, a vector of $d$
numerical values. Let $X\in\R^{L\times d}$ collect these vectors for
the original prompt, which has $L$ token positions. MedQA reference
prompts contain questions and answer options drawn from the training
split. Attribution compares the original and reference embeddings
position by position, so both must have the same dimensions. We
truncate longer reference sequences and extend shorter ones with
padding tokens to obtain $B\in\R^{L\times d}$.

The model also receives an attention mask, which marks positions to
use or ignore as padding, and information about token positions. We
retain the original prompt's mask and positions while changing the
embeddings. Model weights are fixed and dropout is disabled. Write
$F_c(Z)$ for the selected answer's width at embeddings $Z$ under these
settings. In particular, $F_c(B)$ evaluates the prepared reference
embeddings under the original prompt's settings. Processing the
reference prompt normally uses its own sequence and settings and may
give a different width. Our reference checks measure this difference.

\paragraph{Calculating attribution.}
Expected Gradients estimates contributions to the width difference
between the original input and the prepared references. For a sampled
reference $B$, it forms intermediate embeddings
$Z_\alpha=B+\alpha(X-B)$. At $\alpha=0$ these are the reference
embeddings; at $\alpha=1$ they are the original input embeddings;
at $\alpha=0.5$ each coordinate is halfway between the two.

At these intermediate inputs, the gradient measures how a small change
in each embedding coordinate affects width. Multiplying these
sensitivities by the corresponding input--reference differences and
averaging gives
\begin{equation}
 \EG(F_c;X)=\E_{B,\alpha}\left[(X-B)\odot
 \left.\nabla_ZF_c(Z)\right|_{Z=B+\alpha(X-B)}\right].
 \label{eq:eg}
\end{equation}
Here $B$ is sampled from the prepared reference collection,
$\alpha$ is sampled uniformly from $[0,1]$, and $\odot$ multiplies
corresponding coordinates. The expectation $\E$ is estimated by
averaging sampled contributions. We use Captum's GradientShap with
no added Gaussian noise. Gradients pass through the transformer,
output layer and probability conversion, so attribution explains the
final width reported by the classifier.

\subsection{Answer-set attribution and token scores}

Direct EG attributes $F_c$. The answer-set route attributes each
corrected probability contributing to width, then adds its coordinate
contributions. Linearity gives
\begin{equation}
 \EG(F_c;X)=\sum_{\substack{S\ni c\\|S|>1}}\EG(m(S);X).
 \label{eq:decomp}
\end{equation}
The equality follows because differentiation and averaging both
preserve addition. When checking agreement, both routes use the same
sampled references and intermediate inputs, so sampling differences
do not obscure the result (Appendix~\ref{app:proofs}).

The calculation produces a contribution $e_{ij}$ for each embedding
coordinate $j$ of token $i$. To rank tokens, we combine each token's
coordinate contributions into a nonnegative magnitude,
$a_i=(\sum_j e_{ij}^2)^{1/2}$. In the answer-set route, contributions
are added across sets before taking this magnitude, allowing positive
and negative contributions to cancel.

Numerical checks use a different summary: the signed sum
$\sum_{i,j}e_{ij}$, which retains positive and negative contributions
to the width difference. This sum is checked against the original
input's width minus the average width of the prepared references.
The nonnegative token scores serve to rank tokens; they are not
expected to sum to that difference.

\section{Mass correction and credal width}
\label{sec:correction}

The classifier converts predicted lower probabilities into intermediate
answer-set values $\widetilde m_S$. To obtain valid probabilities,
it removes negative values, assigns any shortfall to the full answer
set and normalises. For non-empty proper subsets of $\Theta$, define
\begin{equation}
 q_S=\max(0,\widetilde m_S),\qquad
 s=\sum_{\varnothing\ne S\subsetneq\Theta}q_S,\qquad
 r=\max(1-s,0).
 \label{eq:terms}
\end{equation}
The corrected masses are $m(S)=q_S/(s+r)$ and $m(\Theta)=r/(s+r)$.
Their sum is one because $s+r=\max(s,1)>0$.

Let $u_c=\sum_{S\subsetneq\Theta,\,c\in S,\,|S|>1}q_S$ be the
surviving total on proper sets containing $c$ with other options.
Adding the full-set contribution gives the exact width
\begin{equation}
 W_c=\frac{u_c+r}{s+r}
 =\begin{cases}
 u_c+1-s,&s\leq1,\\
 u_c/s,&s>1.
 \end{cases}
 \label{eq:corrected}
\end{equation}
For $s\leq1$, the shortfall contributes to every answer's width.
For $s>1$, all sets share the denominator $s$, so values assigned
to sets excluding $c$ can change its width. For example, suppose $q_{\{B,C\}}=q_{\{A\}}=0.6$ and all other
values are zero. Their total is 1.2, so $W_C=0.6/1.2=0.5$.
Increasing only $q_{\{A\}}$ to 0.9 raises the total to 1.5 and gives
$W_C=0.6/1.5=0.4$. The value assigned to the set containing C has
not changed before normalisation, but its final probability and C's
width have both decreased.

Where derivatives exist and $s>1$, the input gradient is
\begin{equation}
 \nabla_XW_c=\frac{1}{s}\nabla_Xu_c
             -\frac{u_c}{s^2}\nabla_Xs.
 \label{eq:gradient-excess}
\end{equation}
The first term describes changes in values assigned to sets containing
$c$ and other options. The second describes changes in the total used
to rescale them, including values assigned to sets excluding $c$.
Although this total is shared by all options, its coefficient depends
on $u_c$, so the effect need not be equal for every answer.
Appendix~\ref{app:proofs} gives the full gradient, including the
shortfall term and behaviour at boundaries. Both attribution routes
differentiate the corrected probabilities, so separating answer-set
attributions does not remove this dependence. Correction is part of
training and is reevaluated at every sampled intermediate embedding.

\section{Experimental design}
\label{sec:experiments}

\subsection{Dataset and classifiers}

We use four-option MedQA US \citep{jin2021medqa}, with 10,178
training, 1,272 development and 1,273 test questions. Prompts contain
the question, lettered options and an \texttt{Answer:} prefix, truncated
to 512 tokens. SmolLM3-3B and Llama-2-7B are adapted using LoRA
\citep{hu2022lora}, which learns low-rank weight updates, and a trained
classification output layer. The training loss combines the negative log pignistic probability of
the correct answer with a binary cross-entropy term, weighted by 0.1.
The latter indicates whether each answer set contains the correct
answer. Checkpoints are selected by the lowest development-set answer
negative log-likelihood. Appendix~\ref{app:config} gives the training
settings.

We report classification accuracy on the full test set. To assess the
uncertainty being explained, we also measure whether larger widths
rank incorrect predictions above correct predictions, using
error-detection AUROC. We compare this with ranking predictions by
low probability of the selected answer.

\subsection{Token attribution and zero-masking}

The main study uses direct attribution on the first 100 test questions
per model, keeping the originally predicted answer fixed. SmolLM3
uses 256 training reference prompts and 512 EG samples. Llama averages
coordinate-level attributions from three collections, each containing
1,024 reference prompts and using 1,024 samples. Appendix~\ref{app:config}
reports precision settings and implementation checks.

Eligible tokens belong to the question or answer-option text and
contain a letter or digit. Special tokens, padding, formatting labels,
punctuation-only tokens and the answer prefix are excluded. Let
$R_iX$ set token $i$'s embedding to zero, preserving sequence length
and the attention mask. We measure
\begin{equation}
 D_i(x)=|F_c(R_iX)-F_c(X)|,\qquad
 A(x)=D_{i^*}(x)-\frac{1}{20}\sum_{b=1}^{20}D_{j_b}(x),
 \label{eq:effect}
\end{equation}
Here $i^*$ is the highest-ranked eligible token and the $j_b$ are
20 uniformly sampled eligible positions. A positive $A(x)$ means
that zero-masking the highest-ranked token changes width more than
the average change from zero-masking these randomly selected tokens.
Both changes concern the same selected answer.

Each random draw uses the same eligible positions: positions can
repeat, and the highest-ranked token can also be selected. Random
selection is not restricted to tokens with a similar meaning or
position, or to the same question or option section. Controls therefore
match the zero-masking operation and number of tokens, not these other
properties. We report the mean advantage and a 95\% bootstrap interval,
obtained by resampling questions and recalculating the mean 10,000
times. Each question retains its highest-ranked and random-token
results together during resampling.

\subsection{Probability correction}

We evaluate all test questions under six configurations: output-layer-only,
partial and LoRA adaptation of SmolLM3, and Llama LoRA seeds 7, 17 and
23. We record negative intermediate values, the surviving total $s$,
the added shortfall $r$, and the total absolute change from intermediate
values to final masses. Rescaling is active when $s>1+10^{-6}$.

\subsection{Controlled attribution checks}

Synthetic questions describe fictional conditions in paired versions
with and without a distinguishing test result. The 960 examples are
split into 640 training, 160 development and 160 test examples, keeping
condition families within one split. A softmax layer on frozen SmolLM3
representations predicts probabilities for the 15 non-empty answer sets,
without correction. A later dataset audit found an answer-order shortcut
and unjustified target asymmetry. We use this model to test attribution
calculations; Appendix~\ref{app:controlled} describes the data limitations
and revision.

\paragraph{Reference construction.}
For four test pairs, we explain the prompt containing the result using
three alternatives: training reference prompts, a zero-embedding baseline,
and the paired reference prompt without the result. Adding the test result can change the number of tokens and shift later
text to different positions. For the paired reference prompt, we align
matching token sequences so that unchanged text occupies the original
prompt's positions. Fitting the remaining sequences may omit or repeat
reference tokens; unmatched positions receive padding. Attribution
retains the original prompt's attention mask and token positions.
This alignment is additional to the ordinary truncation and padding
used for MedQA reference prompts.

We evaluate the paired reference prompt normally and then evaluate its
prepared embeddings under the attribution settings. If these widths
differ, the attribution calculation no longer explains the width
difference between the two prompts as normally processed.

\paragraph{Numerical accuracy.}
We repeat attribution calculations using different sample counts and
random seeds. Spearman correlation measures agreement in token order,
while overlap measures how many highly ranked tokens are shared.
We also compare the full attribution vectors, retaining their positive
and negative entries. Cosine similarity measures agreement in their
direction, and relative Euclidean distance measures differences in
their values relative to vector size.

A separate check asks whether the contributions add up to the width
difference being explained. The completeness residual is the absolute
gap between their signed sum and the original input's width minus the
average width of the prepared references:
\begin{equation}
 R=\left|\sum_{i,j}\widehat e_{ij}
 -\left(F_c(X)-\frac{1}{M}\sum_{b=1}^{M}F_c(B_b)\right)\right|,
 \label{eq:residual}
\end{equation}
Here $B_b$ are sampled prepared reference embeddings, evaluated with
the same mask and positions used for attribution. The sum retains
positive and negative coordinate contributions rather than the
nonnegative token scores used for ranking. Initial median and maximum
residual thresholds are 0.01 and 0.05; Appendix~\ref{app:config} gives
the vector and ranking criteria.

For further checks, we use Integrated Gradients (IG) in FP32 and
increase the number of intermediate points between each original input
and prepared reference. Unlike EG's sampling over references and
intermediate points together, this evaluates the full transition for
each chosen reference separately. We refine all four aligned-reference
calculations and, for one case, calculations using 128 training
reference prompts. Each reference's residual is checked before
averaging its contributions with those from other references.
Appendix~\ref{app:localisation} reports the one-case clue analysis.

\section{Results}
\label{sec:results}

\subsection{Zero-masking on MedQA}

Zero-masking the highest-ranked token produces a larger absolute width
change than random zero-masking on both models (Table~\ref{tab:effects}).
Mean advantages are 0.0381 for SmolLM3 and 0.0450 for Llama on the
$[0,1]$ width scale; both 95\% intervals exclude zero.
Full-test accuracies are 49.3\% and 45.7\%, respectively. Width ranks
errors with AUROC 0.634 and 0.598, below the 0.680 and 0.671 obtained
using low predicted-answer probability.

\begin{table}[t]
 \centering
 \caption{Mean advantage in absolute width change from zero-masking
 the highest-ranked rather than random eligible tokens
(Equation~\ref{eq:effect}); 100 MedQA questions per model.
 Intervals are question-level 95\% bootstrap intervals.}
 \begin{tabular}{@{}lrrl@{}}
 \toprule
 Model & Questions & Mean advantage & 95\% CI\\
 \midrule
 \EffectRows
 \bottomrule
 \end{tabular}
 \label{tab:effects}
\end{table}

\subsection{Frequency of probability correction}

The surviving total exceeds one on 95.84--99.92\% of test questions
across the six configurations (Appendix~\ref{app:config},
Table~\ref{tab:correction}). Thus width is usually $u_c/s$ at the
original prompt. Every tested prediction contains negative intermediate
values; the mean total absolute correction ranges from 1.23 to 2.81.
Equation~\ref{eq:gradient-excess} shows how the shared total affects
input sensitivity. Its frequency at test prompts does not measure its
contribution along the intermediate embeddings used for attribution.

\subsection{Numerical accuracy}

On the four synthetic cases, increasing the sample count leaves
training-reference token rankings nearly unchanged: median correlation
is 0.9951. However, median residual is 0.0652 and relative vector
difference is 0.1563, exceeding tolerances of 0.01 and 0.10
(Table~\ref{tab:numerics}). Aligned-reference and zero-baseline
calculations also exceed the residual threshold despite high ranking
agreement.

\begin{table}[t]
 \centering
 \caption{Numerical checks on four synthetic cases. Entries are medians.
Rank correlation and relative vector difference compare sampling
budgets; $R$ is the completeness residual (Equation~\ref{eq:residual}).}
 \begin{tabular}{@{}lrrr@{}}
 \toprule
 Reference & Rank $\rho$ & $R$ & Vector difference\\
 \midrule
 \NumericsRows
 \bottomrule
 \end{tabular}
 \label{tab:numerics}
\end{table}

The ranking and numerical checks answer different questions. Token order can remain unchanged even when contribution values change substantially. Here, the high ranking agreement coexists with a sum of contributions that misses the measured width difference by more than the specified tolerance. This motivates refining the calculation
before interpreting its values.

Denser FP32 IG passes the residual and attribution-agreement checks for
all four aligned references, with maximum residual $1.105\times10^{-5}$.
For one case, refining five of 128 training-reference integrals gives
mean and maximum residuals of 0.00401 and 0.02454. This refinement
evaluates complete IG integrals, not a rerun of stochastic EG.

\subsection{Effects of reference preparation}

Let $w_{\rm with}$ and $w_{\rm without}$ be widths for the paired
prompts processed normally, and $w_{\rm aligned}$ the width of the
prepared reference under the attribution settings. Then
\begin{equation}
 \Delta_{\rm natural}=w_{\rm with}-w_{\rm without},\qquad
 \Delta_{\rm aligned}=w_{\rm with}-w_{\rm aligned}.
 \label{eq:endpoints}
\end{equation}
Alignment shifts the width differences by 0.203--0.537
(Table~\ref{tab:endpoints}). In the first case, supplying the test
result reduces width by 0.478592 when both prompts are processed
normally. Using the aligned reference instead gives a small increase
of 0.002640. The signed IG contributions sum to $+0.002651$, closely
accounting for the aligned-reference difference rather than the
original decrease.

The two checks therefore identify different properties: the small
residual shows that the contributions account for the difference
actually supplied to IG, while evaluating the reference before and
after preparation shows that this difference has changed. A direct
evaluation of both reference versions detects this shift without
requiring an attribution calculation.

\begin{table}[t]
 \centering\small
 \caption{Width differences using the original and aligned paired
reference prompts (Equation~\ref{eq:endpoints}). Shift is
$|\Delta_{\rm natural}-\Delta_{\rm aligned}|$; residuals refer
to IG with aligned references. All evaluations use the same FP32 model.}
 \begin{tabular}{@{}lrrrr@{}}
 \toprule
 Case & $\Delta_{\mathrm{natural}}$ &
 $\Delta_{\mathrm{aligned}}$ & Absolute shift & IG residual\\
 \midrule
 \EndpointRows
 \bottomrule
 \end{tabular}
 \label{tab:endpoints}
\end{table}
\section{Discussion and limitations}

\xurs{} identifies tokens that influence reported credal width under
zero-masking. The correction analysis specifies the function being
explained, while the numerical and reference checks address how that
function is attributed to tokens.

Before interpreting highlighted tokens, we evaluate the reference
prompt normally and under the prepared attribution settings to establish
which width difference the explanation concerns. We then check whether
the signed contributions account for that difference and whether their
values agree when more samples or intermediate points are used. Both
checks matter: a small completeness residual constrains the sum but
can conceal errors in individual contributions that cancel. Finally,
zero-masking tests how selected tokens affect width under an explicit
input modification. Appendix~\ref{app:protocol} gives the procedure.

The evaluation covers fixed 100-question MedQA panels and two
four-option classifiers with different attribution budgets. Zero-masking
changes embeddings rather than deleting or rewording text. These
experiments establish sensitivity to that operation, not clinical
evidence recovery or an independent validation of width as epistemic
uncertainty. The numerical and reference findings concern four synthetic
pairs. Their dataset permits shortcuts, so attribution to the constructed
clue cannot be assessed as though the model necessarily relies on it.
Some follow-up checks were developed after inspecting test results.

Further evaluation should use fresh questions, controlled evidence edits
and meaning-preserving rewordings. Comparing width attribution with
entropy or answer-score attribution under the same references and
computational budgets would test the additional value of explaining
the random-set output. Clinical interpretation would additionally require
domain-expert evaluation.

\section{Conclusion}

\xurs{} attributes an answer's credal width to input tokens. On MedQA,
its highest-ranked tokens produce larger zero-masking effects than
random controls. The correction analysis explains the width being
attributed, while controlled tests show why numerical accuracy and
preservation of the reference prompt require separate checks. Together,
these provide a framework for inspecting the input dependence of
random-set uncertainty estimates.

\label{end:main}
%\clearpage
%\input{sections/08-statements.tex}
\bibliography{references}
\bibliographystyle{iclr2027_conference}
\clearpage
\appendix
\section{Mathematical details}
\label{app:proofs}

This appendix explains the conversion from predicted lower
probabilities to answer-set masses, derives the corrected width,
and establishes the relationship between the two attribution routes.

\subsection{Conversion from lower probabilities to masses}

Let $\Theta$ contain the four answer options.
For each non-empty proper subset $S$ of $\Theta$, the output layer
produces a sigmoid value $b_S$, intended to estimate the lower
probability of $S$. There are 14 such subsets.

A lower probability includes the masses assigned to all subsets
of $S$. Recovering the mass assigned to $S$ therefore involves
subtracting the amounts already assigned to its smaller subsets.
The implementation processes singletons first, followed by pairs
and then triples:
\[
 \widetilde m_{\{c\}}=b_{\{c\}},\qquad
 \widetilde m_S
 =b_S-\sum_{\varnothing\ne T\subsetneq S}
          \max(\widetilde m_T,0),
 \qquad 2\leq |S|\leq3.
\]
Here $\widetilde m_S$ is an intermediate value, not necessarily a
valid probability. All strict subsets of $S$ have been processed
before this expression is evaluated.

The positive-part operation inside the sum is important.
Ordinary M\"obius inversion, the exact algebraic inverse of the
belief-to-mass relationship, would subtract the smaller-set
values with their signs retained. The implemented conversion
instead subtracts only their nonnegative parts.
The subsequent correction removes any remaining negative
values and produces masses summing to one.
The corrected-width derivation below concerns this final
correction and does not assume that the intermediate values
are valid probabilities.

\subsection{Corrected width and its gradient}

After conversion, define
\[
 q_S=\max(\widetilde m_S,0),\qquad
 s=\sum_{\varnothing\ne S\subsetneq\Theta}q_S,\qquad
 r=\max(1-s,0).
\]
Thus $q_S$ is the surviving nonnegative value, $s$ is their total,
and $r$ fills any shortfall below one.
The corrected masses are
\[
 m_S=\frac{q_S}{s+r}
 \quad (\varnothing\ne S\subsetneq\Theta),
 \qquad
 m_\Theta=\frac{r}{s+r}.
\]

For a selected answer $c$, let
\[
 u_c=
 \sum_{\substack{S\subsetneq\Theta\\c\in S,\;|S|>1}}q_S.
\]
Only non-singleton sets containing $c$ contribute to its width.
The proper subsets contribute $u_c/(s+r)$, and the full answer
set contributes $r/(s+r)$. Consequently,
\[
 W_c=\frac{u_c+r}{s+r}.
\]
When $s\leq1$, the added amount is $r=1-s$ and the denominator is
one. When $s>1$, no additional mass is needed and the denominator
is $s$. Hence
\[
 W_c=
 \begin{cases}
 u_c+1-s, & s\leq1,\\[2pt]
 u_c/s, & s>1.
 \end{cases}
\]

To differentiate this expression with respect to the input
embeddings $X$, write $n=u_c+r$ and $d=s+r$.
The quotient rule gives
\begin{align*}
 \nabla_X W_c
 &=
 \frac{\nabla_Xu_c+\nabla_Xr}{d}
 -
 \frac{n}{d^2}\bigl(\nabla_Xs+\nabla_Xr\bigr)\\
 &=
 \frac{1}{d}\nabla_Xu_c
 +
 \frac{s-u_c}{d^2}\nabla_Xr
 -
 \frac{n}{d^2}\nabla_Xs.
\end{align*}
The three terms describe different contributions to the
derivative of the same width, not separate uncertainty measures.

The width formula remains valid when an intermediate value is
zero or when $s=1$. At such boundaries, an ordinary derivative
may not exist; the implementation follows the derivative rules
of its automatic differentiation operations.
During attribution, $s$ and $r$ must be evaluated at each
intermediate embedding between the original prompt's embeddings and the prepared reference embeddings.
Rescaling at the original input does not imply that rescaling
is active throughout that calculation.

\subsection{Equivalence of the attribution routes}

For fixed model weights and a fixed answer $c$, width is a finite
sum of corrected masses:
\[
 F_c(X)=
 \sum_{\substack{S\subseteq\Theta\\c\in S,\;|S|>1}}m_S(X).
\]
Differentiating this sum gives the sum of the derivatives.
Expected Gradients then multiplies by the same input--reference
embedding difference and averages over references and
intermediate points. These operations preserve addition, giving
\[
 \EG(F_c;X)=
 \sum_{\substack{S\subseteq\Theta\\c\in S,\;|S|>1}}
 \EG(m_S;X),
\]
provided the expectations exist.

With the same sampled references and intermediate points, the
finite estimates should therefore agree up to numerical
rounding. Different samples can introduce differences even when
both implementations are correct.

This equality applies to the attribution values before they
are converted into token magnitudes.
Contributions from different answer sets must first be added
with their positive and negative signs retained.
Adding their magnitudes instead would remove cancellation
between contributions and would generally give a different
result.

\subsection{Answer probabilities do not determine credal width}

Consider two valid random-set predictions over four answers.
The first assigns all probability to the full answer set:
$m(\Theta)=1$.
Each answer then has lower probability zero, upper probability
one, and width one.

The second assigns probability $1/4$ separately to each
singleton answer set.
Each answer then has lower and upper probability $1/4$, giving
width zero.

Both predictions produce the same pignistic answer probabilities:
$1/4$ for every option.
A loss that depends only on those answer probabilities cannot
distinguish these two assignments at that input.
This example does not establish that answer-label training
cannot learn useful uncertainty. In particular, our training
objective also contains an answer-set membership loss, described
below, so it is not determined solely by pignistic probabilities.

\section{Model training and attribution settings}
\label{app:config}

\subsection{Classifier training}

The pretrained models are
\path{HuggingFaceTB/SmolLM3-3B} and
\path{meta-llama/Llama-2-7b-hf}.
The LoRA runs begin with previously trained classification
output layers, rather than newly initialised output layers.
These adaptation runs should therefore not be interpreted as
independent training runs from scratch.

The training objective encourages the correct answer to receive
high probability and the predicted lower probabilities to agree
with answer-set membership.
For a correct answer $y$, define $t_S=1$ when $y\in S$ and zero
otherwise. The auxiliary loss is
\[
 \ell_{\mathrm{set}}
 =
 -\frac{1}{14}
 \sum_{\varnothing\ne S\subsetneq\Theta}
 \left[
 t_S\log b_S+(1-t_S)\log(1-b_S)
 \right].
\]
The complete objective is
\[
 \mathcal L
 =
 -\log\BetP(y)+0.1\,\ell_{\mathrm{set}}.
\]
The answer probability $\BetP(y)$ is calculated from the
corrected masses. The conversion and correction operations
therefore participate in training.

\begin{table}[htbp]
 \centering\small
 \caption{Training settings for the main LoRA classifiers.
 Patience is the number of epochs without improvement allowed
 before stopping. Checkpoints are selected using development-set
 answer negative log-likelihood.}
 \begin{tabular}{@{}lll@{}}
 \toprule
 Setting & SmolLM3-3B & Llama-2-7B\\
 \midrule
 LoRA rank & 16 & 8\\
 LoRA scaling parameter & 32 & 16\\
 LoRA dropout & 0.05 & 0.05\\
 LoRA learning rate & $10^{-4}$ & $10^{-4}$\\
 Output-layer learning rate & $5\times10^{-5}$ & $10^{-4}$\\
 Weight decay & 0.01 & 0.1\\
 Warmup fraction & 0.1 & 0.05\\
 Maximum epochs & 5 & 5\\
 Early-stopping patience & 2 & 3\\
 Primary training seed & 7 & 7\\
 Additional correction-study seeds & --- & 17, 23\\
 \bottomrule
 \end{tabular}
 \label{tab:training-settings}
\end{table}

LoRA updates the attention query, key, value, and output
projections, together with the feed-forward gate, up, and down
projections.
Training uses eight workers, each processing one example at a
time and accumulating gradients over four batches before an
update. This gives an effective batch size of 32.
The gradient norm is clipped at one.

\subsection{Expected Gradients settings}

An EG sample selects a reference prompt and an intermediate point
between its prepared embeddings and the original prompt's embeddings.
Table~\ref{tab:attribution-settings} distinguishes the number
of available reference prompts from the number of samples
used in the calculation.

\begin{table}[htbp]
 \centering\small
 \caption{Settings for the main MedQA attribution runs.
 FP16 and BF16 are different 16-bit floating-point formats.
 The samples-per-call setting controls how many samples are
 processed together.}
 \begin{tabular}{@{}lll@{}}
 \toprule
 Setting & SmolLM3-3B & Llama-2-7B\\
 \midrule
 Reference collections & 1 & 3\\
 Prompts per collection & 256 & 1,024\\
 EG samples per collection & 512 & 1,024\\
 Attribution base seed & 11 & 11\\
 Samples processed per call & 8 & 2\\
 Model arithmetic & FP16 & BF16\\
 Averaging across collections & Not applicable & FP32\\
 \bottomrule
 \end{tabular}
 \label{tab:attribution-settings}
\end{table}

SmolLM3 uses the first 256 training prompts as references.
For Llama, training examples are placed in a reproducible
order using seeded SHA-256 hash ranking with seed 20260804.
The first 3,072 examples in this order form three consecutive,
non-overlapping collections of 1,024 references.
The three attribution vectors are averaged with equal weights
in FP32 before token magnitudes are calculated.

\subsection{Checks of Llama ranking agreement}

The final Llama settings are checked on 20 selected questions
from the 100-question panel.
Attributions for the predicted answer and correct answer are
assessed separately.
We compare the average from all three reference collections
with averages omitting each collection in turn, and repeat the
three-collection calculation with attribution seeds 17 and 23.
The comparisons that omit one collection reuse calculations
from the full average; they are not independent replications.

For section-level comparisons, token magnitudes are summed
within the question, each of the four options, and the
\texttt{Answer:} prefix.
Each sum is expressed as a share of total magnitude.
We then compare the ordering of these sections and the
differences between their shares.

\begin{table}[htbp]
 \centering\small
 \caption{Criteria used for the final Llama ranking checks.
 These assess agreement between repeated calculations, not
 whether the attributions explain meaningful evidence.}
 \begin{tabular}{@{}lr@{}}
 \toprule
 Measurement & Required value\\
 \midrule
 Median Spearman correlation between section rankings
   & $\geq0.90$\\
 Fraction agreeing on the highest-attributed option
   & $\geq0.80$\\
 Mean overlap among the 20 highest-ranked tokens
   & $\geq0.75$\\
 Median mean absolute difference between section shares
   & $\leq0.02$\\
 Maximum difference in the selected answer's width
   & $\leq10^{-6}$\\
 \bottomrule
 \end{tabular}
 \label{tab:llama-ranking-checks}
\end{table}

The highest-attributed option is the option whose text receives
the largest attribution share, not necessarily the option
predicted by the classifier.
Token overlap is the fraction of selected token positions
shared by the two rankings.
These ranking checks are separate from the completeness
checks below.

\subsection{Checks of attribution values}

For the four synthetic examples, we compare both token rankings
and attribution values.
The latter comparisons retain the positive and negative
contributions for individual embedding coordinates, before
conversion into token magnitudes.

For two nonzero attribution vectors $e$ and $e'$, cosine
similarity measures agreement in their pattern of contributions,
allowing an overall scale difference.
Relative Euclidean distance measures the size of their
difference relative to the second vector:
\[
 \operatorname{cosine}(e,e')
 =
 \frac{e^\top e'}{\|e\|_2\|e'\|_2},
 \qquad
 \operatorname{relative\ distance}(e,e')
 =
 \frac{\|e-e'\|_2}{\|e'\|_2}.
\]
Completeness is assessed using the residual $R$ in
Equation~\ref{eq:residual}.

\begin{table}[htbp]
 \centering\small
 \caption{Criteria for the initial four-example numerical study.
 Agreement is assessed between attribution calculations using
different sample counts and random seeds for the same original
prompt and reference collection.}
 \begin{tabular}{@{}lr@{}}
 \toprule
 Measurement & Required value\\
 \midrule
 Median completeness residual & $\leq0.01$\\
 Maximum completeness residual & $\leq0.05$\\
 Median Spearman correlation between token rankings
   & $\geq0.95$\\
 Median overlap among the 20 highest-ranked tokens
   & $\geq0.80$\\
 Median cosine similarity between attribution vectors
   & $\geq0.99$\\
 Median relative Euclidean distance between vectors
   & $\leq0.10$\\
 \bottomrule
 \end{tabular}
 \label{tab:controlled-numerical-checks}
\end{table}

The later FP32 Integrated Gradients calculations use a mean
per-reference residual threshold of 0.01 and a maximum of 0.05.
Agreement between integration resolutions requires cosine
similarity of at least 0.99, relative Euclidean distance of at
most 0.10, token-ranking correlation of at least 0.95, and
top-three overlap of at least 0.80.
For the aligned paired references, these ranking comparisons
concern the nine positions changed by alignment.

For complete IG integrals, residuals are checked separately for each
prepared reference before averaging, preventing errors of opposite
signs from cancelling unnoticed. A single EG sample evaluates only
one intermediate point and is not required to account for the full
input--reference width difference.

These thresholds are operational choices for the reported
studies, not universal guarantees of attribution quality.

\subsection{Computing environment}

Early experiments used an NVIDIA T4.
Later experiments used Quadro RTX 6000 and NVIDIA A30 devices
with 24 GiB of memory per GPU.
Independent attribution jobs were distributed across devices;
their memory was not combined to process a single explanation.

Hardware and software environments differed across runs.
The environment used for the later controlled-model study
should not be treated as the environment for every MedQA run.
We do not report a controlled comparison of end-to-end runtime
or energy consumption.

\subsection{Supplementary implementation and correction results}

In a constructed model with one deliberately influential token, both
attribution routes rank that token first. For a separate SmolLM3
classifier with frozen transformer weights, direct and answer-set
attributions have minimum signed-vector cosine similarity 0.999986
and maximum relative Euclidean distance 0.00522. The Llama check
compares the scalar width with the sum of its contributing answer-set
probabilities, with maximum difference $2.98\times10^{-8}$; it is
not a comparison of attribution vectors.

One Llama attribution calculation gives non-finite values in FP16
and finite values in BF16. A single collection of 1,024 reference
prompts with 1,024 EG samples fails the ranking checks. Averaging
coordinate-level attributions across three collections meets the
recorded criteria. Leave-one-collection-out averages share components,
so these checks are not independent ensemble replications.

Table~\ref{tab:correction} reports correction measurements for every
training configuration. The intermediate vectors can contain negative
values and are not probability distributions; their total absolute
adjustment is therefore not a distance between two valid probability
distributions.

\begin{table}[t]
 \centering
\caption{Correction on 1,273 test questions per configuration.
Rescaling means $s>1+10^{-6}$. Adjustment is the mean total absolute
change from intermediate values to final masses. Negative values
occur in every tested prediction.}
 \begin{tabular}{@{}lrr@{}}
 \toprule
 Configuration & Rescaling (\%) & Mean total adjustment\\
 \midrule
 \CorrectionRows
 \bottomrule
 \end{tabular}
 \label{tab:correction}
\end{table}

\section{Attribution to the distinguishing test result}
\label{app:localisation}

This analysis examines one synthetic question,
\texttt{family007\_case00\_resolved}, whose text includes a test
result distinguishing between two plausible answers.
The clue occupies eight tokens.
We ask how much attribution magnitude it receives relative to
other text of the same length.

\subsection{Reference prompts and measurements}

We use two IG calculations that passed the numerical checks. The
first averages attributions over 128 training reference prompts.
The second uses the paired reference prompt without the distinguishing
test result, after alignment. They explain width differences relative
to different reference embeddings.

For each calculation, we examine two regions:
all 205 active, non-special text tokens, and the 29 tokens
within the patient case description.
The full-text region includes the instructions, fictional
diagnostic rules, and answer options.
It is not restricted by the token-eligibility rule used in
the MedQA zero-masking experiment.

The clue's magnitude share is the sum of its token magnitudes
divided by the sum for the selected region.
Control windows contain the same number of tokens as the clue,
are contiguous within that region, and exclude clue tokens.
We compare the clue with the mean magnitude of these windows.

\begin{table}[htbp]
 \centering\small
 \caption{Attribution to the distinguishing clue in one example.
 ``Difference'' is the clue's magnitude share minus the mean
 control-window share, expressed in percentage points.
 ``Best rank'' is the rank of the highest-ranked clue token
 within the selected region; rank one is highest.}
 \begin{tabular}{@{}llrrr@{}}
 \toprule
 IG reference & Region & Clue share (\%) &
 Difference (pp) & Best rank\\
 \midrule
 Training prompts & All text & 2.07 & $-1.33$ & 44\\
 Training prompts & Case text & 22.97 & $-3.40$ & 5\\
 Aligned paired prompt & All text & 52.59 & $+50.89$ & 1\\
 Aligned paired prompt & Case text & 52.59 & $+33.02$ & 1\\
 \bottomrule
 \end{tabular}
 \label{tab:localisation}
\end{table}

With training prompts as references, the clue receives 2.07\%
of magnitude across the full text, or 22.97\% when the
denominator includes only the case description.
Its full-text magnitude is at approximately the 40th percentile
of the equal-length control windows.
The windows overlap, so they are not independent observations
and this percentile is descriptive.

IG using the aligned paired reference assigns 52.59\% of magnitude
to the clue and ranks one of its tokens first. This reference changes
the width difference (Table~\ref{tab:endpoints}) and gives zero
attribution to tokens with embeddings identical to the original input.
The larger share therefore reflects a different attribution task,
not an isolated improvement in clue recovery.

Magnitude shares are not signed fractions of the width change.
The clue is also known from the question's construction, not
from evidence that the trained model relies on it.
The dataset limitations in Appendix~\ref{app:controlled}
therefore restrict the interpretation of this example.

\subsection{Number of integration points}

Integrated Gradients approximates its integral by evaluating
gradients at intermediate points between the original prompt's embeddings and the prepared reference embeddings.
The final training-reference calculation uses 512 points for
123 references and 2,048 points for the remaining five:
\[
 123\times512+5\times2048=73{,}216
\]
gradient evaluation points.

Using 2,048 points for every reference would require
\[
 128\times2048=262{,}144
\]
points. The final calculation therefore contains 27.9\% of
that nominal total.
However, the uniform 2,048-point calculation was not run,
so its accuracy is unknown.

These totals count only the points contributing to the final
estimate. Earlier calculations, repeated evaluations, and
diagnostic checks required additional computation.
The comparison describes how integration points were allocated;
it is not a measured end-to-end speedup.

\section{Synthetic dataset revisions and model evaluation}
\label{app:controlled}

This appendix explains why the original synthetic data can
support checks of attribution calculations but cannot establish
reliable recovery of meaningful evidence.
It also describes a revised dataset and the limits of the
models trained on it.
We refer to the original and revised datasets as V1 and V2.

\subsection{Limitations of the original dataset}

Against its generated labels, the original controlled model
achieves approximately 91.7\% mean test accuracy.
The mean Jensen--Shannon divergence between predicted and
prescribed answer-set probabilities is 0.0208; this measures
their disagreement, with zero indicating equality.
The mean absolute width error is 0.0502.
Across three training seeds, every evaluated question pair
has lower width after the distinguishing result is supplied.

Subsequent inspection found two problems.
First, the diagnostic test named first in the case description
identified the labelled answer with 100\% accuracy.
This made it possible to predict the label without correctly
interpreting which result belonged to which test.
Second, questions without distinguishing results assigned
unequal singleton probabilities to the two plausible answers,
favouring the labelled answer without justification in the text.

We did not establish whether the trained model actually used
the ordering shortcut. Its availability, together with the
unsupported target preference, is enough to prevent treating
high accuracy as evidence that the model learned the intended
reasoning.
The model remains usable for checking whether attribution
values account for a specified output difference.

\subsection{Construction of the revised dataset}

V2 addresses the identified problems in several ways:
\begin{itemize}
 \item The order of diagnostic tests is varied independently
 of the correct answer.
 \item Each example appears in four cyclic arrangements of
 the answer options, with the probability targets remapped
 to preserve the underlying answer identities.
 \item When the text does not distinguish the two plausible
 answers, their singleton target probabilities are equal.
 \item Some paired inputs change only a detail that does not
 affect the prescribed answer-set probabilities, such as
 whether the visit occurs in the morning or evening.
 \item Fictional condition families are separated across
 training, development, and test data. Development and test
 data also include wording templates absent from training.
\end{itemize}

The revised dataset contains 7,680 training, 3,840 development,
and 3,840 test records, drawn from 80, 20, and 20 condition
families respectively.
Different option arrangements and wording variants of a case
are related observations, not independent examples.

A rule-based reader checks that each generated prompt expresses
the intended test results and matches its prescribed target.
We also check that paired prompts can be tokenised with the
same length and attention mask without deleting, duplicating,
or aligning tokens.
These are checks of the generated data and its representation,
not tests of whether a trained model understands it.

Additional simple predictors examine whether labels can be
predicted from test order or from text with the test results
hidden. They are fitted using training data only.
These diagnostics have no pass/fail threshold and do not prove
that every possible shortcut has been removed.

\subsection{Evaluation before attribution}

Before explaining a V2 model, we require it to reproduce the
prescribed probabilities and respond appropriately to changes
in the text.
For probability assignments $\widehat m$ and $m$, the total
absolute error is
\[
 E_m=\sum_{\varnothing\ne S\subseteq\Theta}
       |\widehat m_S-m_S|.
\]
Width error is the mean absolute difference between predicted
and prescribed widths across the four answers.

The evaluation checks the following:
\begin{itemize}
 \item Mean total probability error is at most 0.20, assessed
 separately for questions with and without a distinguishing
 result.
 \item Mean width error is at most 0.05, and accuracy on
 questions with a distinguishing result is at least 0.80.
 \item The selected answer's width is lower for the prompt containing the
distinguishing result than for its paired prompt without that result
in at least 90\% of pairs.
 \item Changing only an irrelevant detail produces mean total
 probability change at most 0.05 and mean absolute width
 change at most 0.02.
 \item Rearranging answer options produces mean total
 probability change at most 0.10 after predictions are mapped
 back to the same answer identities.
\end{itemize}

Predictions must also be nonnegative and sum to one within
$10^{-6}$.
The behavioural criteria are applied separately to familiar
and previously unseen wording templates.
Results are averaged within condition families before
aggregation, so repeated presentations are not treated as
independent cases.

\subsection{Training results on the revised dataset}

We fit linear output layers and nonlinear output layers with
128 hidden units to frozen SmolLM3 representations, using three
training seeds.
The nonlinear model completes its 100-epoch schedule by
continuing from the saved training state.

Table~\ref{tab:controlled-v2-fit} shows the final total
probability errors for that model.
Training errors are substantially smaller than development
errors on previously unseen wording.

\begin{table}[htbp]
 \centering
 \caption{Mean total absolute error in the predicted answer-set
 probabilities after the nonlinear model's 100-epoch schedule.
 Development values concern wording templates not used
 during training.}
 \begin{tabular}{@{}lrr@{}}
 \toprule
 Training seed & Training error & Development error\\
 \midrule
 7  & 0.1221 & 0.4263\\
 17 & 0.1176 & 0.4261\\
 23 & 0.1173 & 0.4359\\
 \bottomrule
 \end{tabular}
 \label{tab:controlled-v2-fit}
\end{table}

No evaluated development checkpoint meets all the required
criteria. We therefore report no V2 attribution experiment.
This result concerns the tested models and training procedure;
it does not establish that random-set uncertainty cannot be
learned.

\section{Reference preparation and numerical checks.}
\label{app:protocol}

The following procedure separates changes caused by reference
preparation from errors in the attribution calculation.

\begin{enumerate}
 \item \textbf{Define the width difference.}
 Choose the answer option and fix the model weights and precision.
 Evaluate the original prompt and intended reference prompt with
 their normal tokenisation, masks and positions; record both widths.

 \item \textbf{Evaluate the prepared reference.}
 Record the prepared reference embeddings and their token identities.
 Evaluate them using the exact mask and positions used for attribution.

 \item \textbf{Check the effect of preparation.}
 Calculate the original prompt's width minus each of the two reference
 widths. Report their difference and signs, as in
 Equation~\ref{eq:endpoints}. Inspect token changes as well: equal
 widths alone do not establish that prompts convey the same information.

 \item \textbf{Check numerical accuracy.}
 Compare the signed attribution sum with the width difference for the
 prepared reference. Repeat with more EG samples or IG integration
 points, checking attribution vectors as well as rankings. For complete
 IG integrals, inspect residuals for individual references before
 averaging; a single EG sample need not satisfy completeness.

 \item \textbf{Evaluate highlighted evidence.}
 State the eligible token region and any clue known from the task's
 construction. Compare its attribution magnitude with that of
 equal-length control windows. Record positions forced to zero by
 identical input and reference embeddings, and account for related
 versions of the same question when aggregating results.
\end{enumerate}

The procedure checks the reference and numerical calculation. Whether
the learned width represents epistemic uncertainty and whether
highlighted tokens identify meaningful evidence require separate
model and task evaluations.
\end{document}